\documentclass[11pt]{article}
\usepackage[utf8]{inputenc}
\usepackage{graphicx}
\usepackage{booktabs}
\usepackage{array}
\usepackage{tabularx}
\usepackage{amsmath}
\usepackage{amssymb}
\usepackage{listings}
\usepackage{xcolor}
\usepackage[hidelinks]{hyperref}
\usepackage{seqsplit}
\usepackage{placeins}
\usepackage[margin=1in]{geometry}
\title{A doctrine-grounded visual question answering dataset for Tactical Combat Casualty Care}
\author{Junseob Kim\,$^{1,*}$, Jade Chng\,$^{2}$, Ayman Ali\,$^{3}$, Victor Moas\,$^{3}$,\\
Yichun Lee\,$^{4}$, Po-Chun Chin\,$^{4}$, Sunil Hwang\,$^{5}$, Rishikesan Kamaleswaran\,$^{3,6}$\\[4pt]
{\small $^{1}$Department of Electrical and Computer Engineering, Duke University, Durham, NC, USA}\\
{\small $^{2}$Department of Biomedical Engineering, Duke University, Durham, NC, USA}\\
{\small $^{3}$Department of Surgery, Duke University, Durham, NC, USA}\\
{\small $^{4}$School of Medicine, College of Medicine, National Yang Ming Chiao Tung University, Taipei, Taiwan}\\
{\small $^{5}$Department of Mathematics, Korea Military Academy, Seoul, South Korea}\\
{\small $^{6}$Department of Anesthesiology, Duke University, Durham, NC, USA}\\[2pt]
{\small $^{*}$Corresponding author: Junseob Kim (junseob.kim@duke.edu)}}

\date{}

\begin{document}
\maketitle

\begin{abstract}
Tactical Combat Casualty Care (TC3) requires responders to connect visual observations of injuries and interventions with established clinical guidance. Developing vision-language models to support this process requires supervision that links visible evidence to traceable doctrine. We present TC3-VQA, a dataset constructed from public instructional and field TC3 videos and authoritative TC3 documents. It contains 581 items spanning 11 concepts, with 1,860 questions covering intervention recognition, doctrine, clinical reasoning, procedural guidance, and refusal when visual information is insufficient. Doctrine-based answers preserve verbatim source passages and character offsets. Construction combines visual annotation, passage retrieval, entailment checks, and verification across model families. Equipment boxes, anatomical labels, temporal segments, and source metadata accompany the question-answer pairs. Automated audits and ratings by two physicians and two medical students characterize annotation quality, with human ratings available for 88 retained items. The dataset provides a resource for adapting vision-language models to TC3, studying the connection between visual evidence and clinical knowledge, and evaluating recognition, doctrine recall, and abstention.

\end{abstract}

\begin{figure*}[t]
\centering
\includegraphics[width=\textwidth]{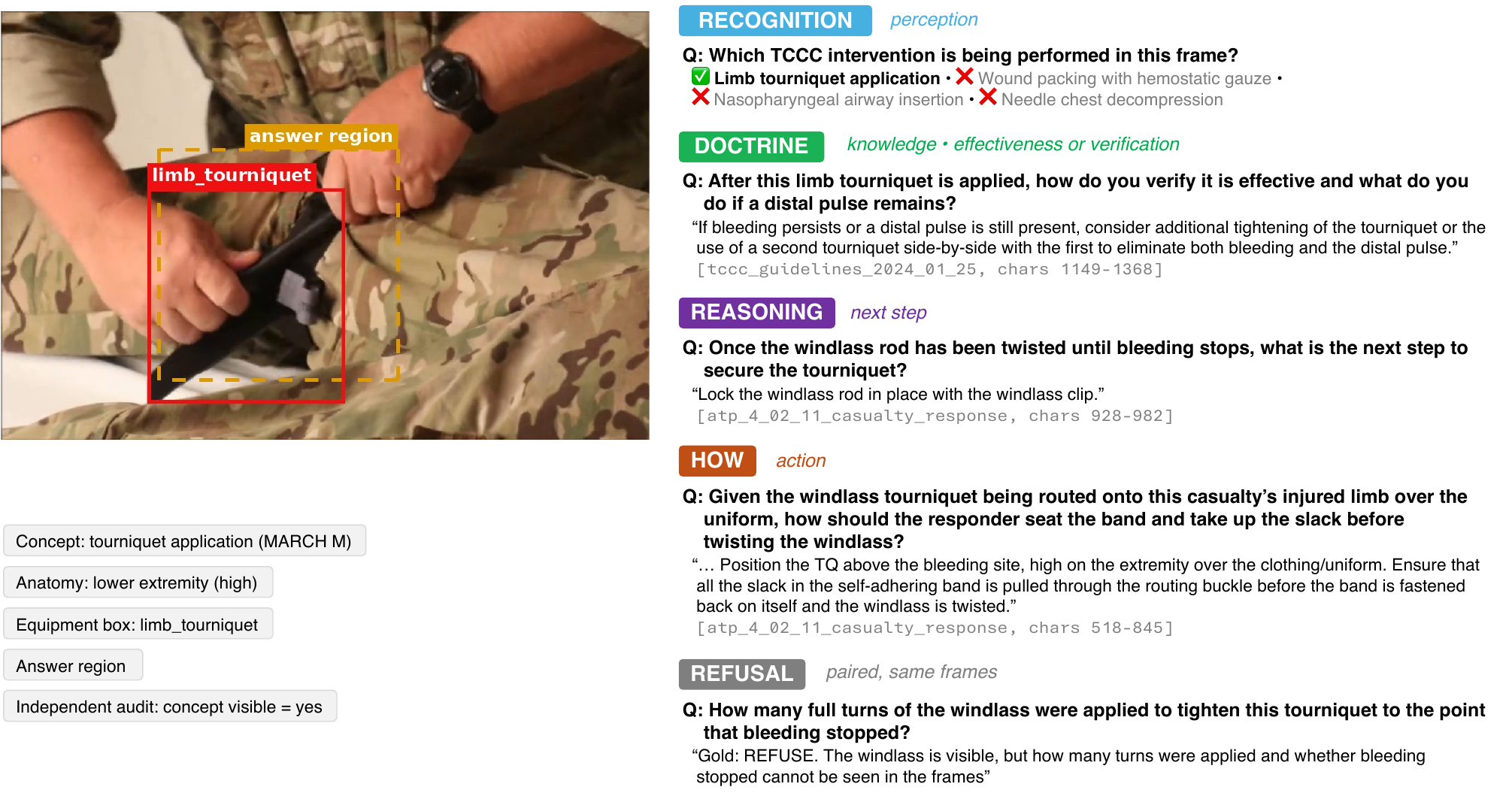}
\caption{Example TC3-VQA item (\texttt{ans\_01332}) from public-domain footage of the Committee on Tactical Combat Casualty Care (CoTCCC). Left: one of four frames with equipment, answer-region, and anatomy annotations. Right: four answerable question types with cited doctrine spans. The paired refusal question asks for the number of windlass turns, which the available frames do not show.}
\label{fig:example}
\end{figure*}

\section{Background \& Summary}
\label{sec:intro}

Tactical Combat Casualty Care (TC3) provides guidance for treating battlefield injuries before definitive medical care is available~\cite{cotccc2024guidelines,army2022atp40211}. Among U.S. combat fatalities in Iraq and Afghanistan from 2001 to 2011, $87.3\%$ occurred before arrival at a medical treatment facility~\cite{eastridge2012death}. When evacuation is delayed, care may extend well beyond the initial response, placing sustained demands on limited personnel and equipment~\cite{keenan2017prolonged}. Responders can vary in clinical experience, and combat lifesaving skills can decline after initial training~\cite{nemeth2021decision,landman2023retention}.

Computational approaches to prehospital trauma decision support have largely focused on physiological monitoring. Machine-learning systems use vital signs and physiological waveforms to assess shock, hemorrhage risk, and the need for life-saving interventions~\cite{nemeth2021decision,nemeth2023tccc,convertino2020wearable,liu2014lsi}. The DARPA Triage Challenge further reflects broader interest in technology-assisted casualty assessment under resource constraints~\cite{darpatriage2024}. Physiological signals, however, do not directly show which equipment is present or how an intervention is being performed. Vision-language models (VLMs) offer a complementary approach by connecting visual observations of equipment and interventions to natural-language questions and guidance. TC3-VLM demonstrated the feasibility of adapting general VLMs to TC3 videos using visually grounded and caption-derived question-answer pairs~\cite{kim2026tc3vlm}. Supporting knowledge-intensive questions in this setting requires training data that connect visible interventions to verifiable clinical knowledge.

Medical visual question answering (VQA) provides a practical framework for learning these connections, with potential applications in clinical decision support~\cite{lau2018vqarad}. Datasets including VQA-RAD, PathVQA, SLAKE, and PMC-VQA cover radiology, pathology, and other medical imagery~\cite{lau2018vqarad,he2020pathvqa,liu2021slake,zhang2024pmcvqa}. More recent efforts have increased scale by generating instruction data from biomedical figure captions, narrated histopathology videos, and public educational videos~\cite{li2023llavamed,seyfioglu2024quiltllava,openbiomedvid}. Trauma THOMPSON extends this direction to egocentric simulated trauma procedures annotated for actions, hands, tools, and medical VQA~\cite{zhuo2026traumathompson}. However, comparable resources for TC3 remain limited, particularly for questions that connect visible battlefield-care procedures with relevant clinical guidance.

Constructing such a resource from public visual material introduces challenges in both visual and knowledge grounding. General and medical VLMs are susceptible to visual hallucinations~\cite{li2023pope,gu2024medvh}, so using the same model to interpret a scene and write its reference answer can propagate an initial error into the training target. Video narration can also be temporally misaligned with the corresponding image~\cite{han2022tan}, while language priors can enable some questions to be answered without using the image~\cite{chen2024mmstar}. Knowledge grounding likewise requires explicit verification: SourceCheckup found that cited references often did not substantiate every claim in generated medical responses~\cite{wu2025sourcecheckup}. These limitations motivate dataset construction that independently verifies what the scene supports and what the clinical source substantiates.

We therefore present TC3-VQA, a dataset linking public TC3 footage to authoritative clinical guidance. It contains $1{,}860$ questions on $581$ items ($431$ answerable and $150$ refusal) across $11$ concepts, covering intervention recognition, doctrine, clinical reasoning, procedural guidance, and answerability (Figure~\ref{fig:example}). The release provides traceable source passages alongside equipment boxes, anatomical labels, temporal annotations, source metadata, and quality assessments. TC3-VQA is intended to support VLM fine-tuning and evaluation of visual grounding, doctrine-based reasoning, and appropriate abstention in TC3.

\section{Methods}
\label{sec:benchmark}

\begin{figure*}[t]
\centering
\includegraphics[width=0.95\textwidth]{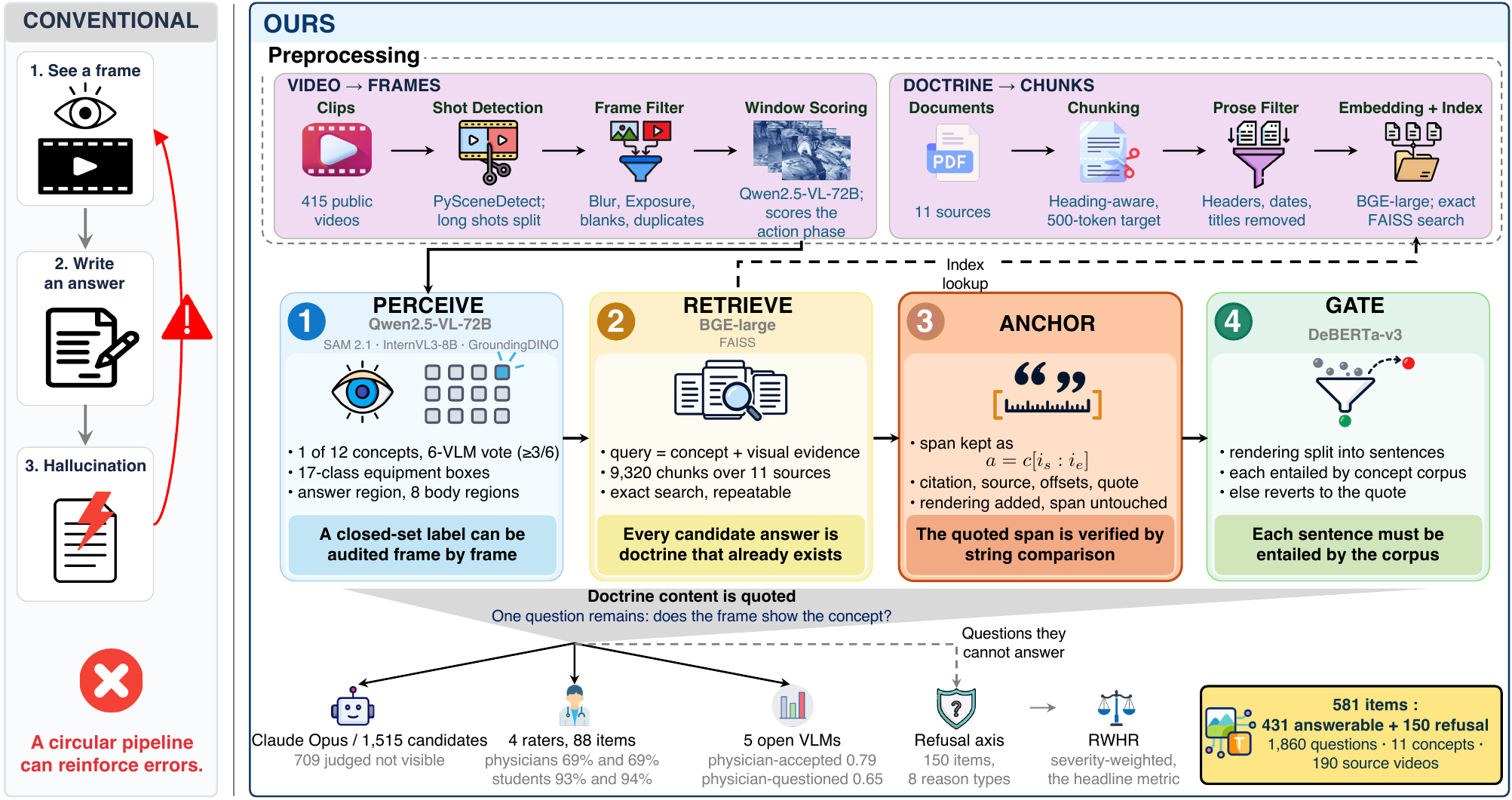}
\caption{TC3-VQA construction workflow. Left: direct generation of questions and answers from an image. Right: visual concept assignment, doctrine retrieval, verbatim answer anchoring, and entailment filtering. Equipment and anatomy annotations record visual evidence; cross-family review and recognition consensus check concept labels. Refusal questions target information absent from the frames.}
\label{fig:pipeline}
\end{figure*}

Figure~\ref{fig:pipeline} provides an overview of the construction workflow.

\subsection{Data collection and preprocessing}
\label{sec:data-sources}

\textbf{Video collection and preprocessing.}
Search terms were developed with a physician trained in TC3 and covered the interventions and phases of care defined in the guidelines of the Committee on Tactical Combat Casualty Care (CoTCCC)~\cite{cotccc2024guidelines}. Using these terms, we collected $415$ publicly available YouTube videos. The collection combined formal instructional material with field footage of casualty care, spanning both procedural demonstrations and interventions performed under realistic conditions. Video identifiers, URLs, and timestamps were retained to preserve source traceability.

\label{sec:frame-extraction}
Videos were split into shots using the PySceneDetect content detector~\cite{pyscenedetect} at its default threshold ($27$). Shots longer than $60$\,s were divided into windows of at most $30$\,s, and frames were sampled approximately every $4$\,s. Filtering for poor exposure, blank content, blur, and near-duplicates left $32{,}987$ frames in $17{,}055$ windows. Because a single frame can be ambiguous about whether an intervention is under way, Qwen2.5-VL-72B-Instruct~\cite{qwen25vl} evaluated each window jointly in temporal order. Each frame received an action phase (active procedure, post-procedure result, assessment only, or no medical action) and a usefulness score from $0$ to $5$. Windows containing an active procedure or post-procedure result with a usefulness score of at least $3$ were retained with up to four qualifying frames, yielding $8{,}012$ windows.

\textbf{Doctrine corpus collection and preprocessing.}
The doctrine corpus contained $9{,}320$ chunks from $11$ sources spanning operational guidance and supporting clinical literature. The CoTCCC guidelines, including the January 2024 edition~\cite{cotccc2024guidelines} and the December 2021 guidelines for medical personnel~\cite{cotccc2021tcccmp}, define the TC3 standard of care. Army Techniques Publication (ATP)~4-02.11~\cite{army2022atp40211} provides procedural guidance, while the Joint Trauma System Clinical Practice Guidelines~\cite{jts2024cpg} add intervention-specific clinical detail. Joint Publication 4-02~\cite{jcs2017jp402} provides the broader joint health-services context. Of these, ATP~4-02.11, the A--H collection of Joint Trauma System guidelines, and the two CoTCCC guideline documents contributed $1{,}902$ chunks covering the interventions represented in the dataset. The remaining corpus broadened retrieval to scene-specific guidance, acronym definitions, and related literature through a glossary compiled from the doctrine documents and PubMed and arXiv abstracts retrieved with TC3-focused queries.

Documents were extracted with PyMuPDF and split at detected headings into chunks of approximately $500$ tokens, with a maximum of $800$, to preserve recommendations together with the sections defining their scope. PubMed and arXiv records and glossary entries were treated as individual chunks. Each chunk retained its source, section path, position, and page range, and duplicates arising from overlapping guideline archives were removed. Before retrieval, a prose filter excluded short, numeric, capitalized, and bibliographic fragments, removing $1{,}428$ of the $5{,}142$ chunks derived from guideline and doctrine documents.

\subsection{Visual annotation and concept assignment}
\label{sec:grounding}

Each retained window received one of $12$ observable TC3 intervention or documentation labels, organized by the MARCH sequence (massive hemorrhage, airway, respiration, circulation, and hypothermia) and a documentation category (Table~\ref{tab:concepts}). Qwen2.5-VL-72B assigned the initial concept, generated a factual scene description, and rated its confidence as high, medium, or low. High- and medium-confidence assignments ($5{,}051$ of $8{,}012$ windows) formed the answerable pool, while low-confidence windows were reserved for refusal construction. Limiting each video to six candidates reduced source concentration and yielded $1{,}515$ items for review.

Because concept errors would propagate to questions derived from a scene, labels underwent independent verification before question generation. Label consistency was first tested under a stricter perception prompt, followed by a blinded visual review with Claude Opus~4.8 that classified each concept as clearly visible, partly visible, or not visible without access to the initial reasoning. Of the $1{,}515$ candidates, $336$ were clearly visible, $470$ partly visible, and $709$ not visible; only the first two groups continued.

Visual annotations provided additional evidence for each retained label. Equipment boxes followed a $17$-class ontology of TC3 instruments and supplies. Initial box proposals were cross-checked for object presence with GroundingDINO~\cite{liu2023groundingdino} and InternVL3-8B~\cite{internvl3}, then refined from object masks using SAM\,2.1 (Hiera-Large)~\cite{ravi2024sam2}. The original box was retained when refinement produced an empty or implausibly large mask. A final visual review checked object presence and class assignment, retaining, relabeling, or removing boxes as needed, with a targeted pass for flagged omissions. Missing boxes therefore do not indicate equipment absence.

The casualty body region was annotated using eight anatomical regions, with additional labels for multiple or non-specific sites and unclear views. An answer-region box marked the visual evidence for the assigned concept. A final consistency check incorporated these annotations to reassess the concept label, with changed assignments reviewed for correction or routing to refusal.

\subsection{Doctrine retrieval and question construction}
\label{sec:corpus-first}
\label{sec:two-axis}

Doctrine was retrieved before question generation so that each answer could be anchored to the source corpus. A query combining the assigned concept with verified visual attributes was embedded with BGE-large-en-v1.5~\cite{baai2023bge}, and FAISS~\cite{douze2024faiss} performed exact inner-product search over L2-normalized vectors. A curated passage for each concept provided an on-topic anchor. To diversify clinical content, doctrine facets such as indications, techniques, key parameters, common errors, and contraindications were assigned to scenes in rotation. Because facet assignment reused the scene description, it was not treated as an independent validation step.

Question construction used the frames and retrieved passages to produce four answerable types. Recognition questions asked which intervention was depicted, using the verified concept as the answer and visually confusable interventions as distractors. Doctrine questions asked for guidance on the depicted intervention and assigned facet, incorporating visible details when supported by the source. Reasoning questions asked for an assessment, next step, or critique of the scene, while procedural \textsc{how} questions asked how to perform, secure, or place the intervention and drew primarily on the operational field manual. Some doctrine questions were intentionally general enough to apply across scenes of the same intervention.

Answers to doctrine, reasoning, and \textsc{how} questions were extracted directly from the retrieved passages. Claude Opus~4.8 selected a verbatim span from the retrieved passages, recorded by corpus chunk and character offsets. A readable rendering was retained only when supported by the entailment check described below; otherwise, the original span was used. Token-preserving normalization removed formatting artifacts while preserving the original span and offsets for provenance.

Refusal questions reused frames from answerable items but requested information not established by the frames, including vital signs, elapsed time, blood-loss volume, off-frame anatomy, illegible medication information, or the outcome of an unshown step. Each question underwent a separate answerability check.

\subsection{Verification and release curation}
\label{sec:verification}
\label{sec:released}

Reference answers were checked for exact source correspondence. For a cited corpus chunk $c$ and recorded character interval $[i_s,i_e)$, a span $a$ was accepted only when
\begin{equation}
a \;=\; c\,[\,i_s : i_e\,].
\label{eq:offset}
\end{equation}
This deterministic comparison verified provenance for every released doctrine, reasoning, and \textsc{how} answer.

A readable rendering $r$ was retained only when every sentence was entailed by the item's concept corpus $C$:
\begin{equation}
\mathrm{accept}(r) \iff \forall\, u \in \mathrm{sent}(r): \; \Pr\nolimits_{\textsc{nli}}\!\left(\text{entail} \mid C, u\right) \ge \tau ,
\label{eq:nli}
\end{equation}
using a DeBERTa-v3 natural language inference (NLI) model~\cite{debertav3,laurer2024nli} with $\tau=0.5$. The same classifier and threshold are used for free-form scoring in the Usage Notes, where entailment must also exceed contradiction.

Each assembled item was reviewed against its frames for concept and anatomy labels, image quality, answerability, answer leakage, doctrine relevance, corpus noise, and answer specificity. Flagged questions were revised under stronger scene constraints and rechecked, reducing off-topic flags from $86$ to $5$, corpus-noise flags from $15$ to $0$, and answer-telegraphing flags from $72$ to $1$. Uncertain anatomy was removed from question stems, answer-revealing captions were masked, and passages inconsistent with the depicted phase of care were replaced or removed. Unusable items were excluded or converted to refusals when possible. Rejected items and audit logs were retained, with potentially unsafe edits flagged for clinician review.

Two final filters determined the released set. Unavailable source videos were first removed together with their $22$ items ($16$ answerable and six refusal), with reasons recorded in the exclusion manifest. The remaining answerable items underwent an independent cross-family consensus check using six VLMs from five families (Qwen2.5-VL-72B, Qwen3-VL-32B~\cite{qwen3vl}, InternVL3-78B, Mistral-Small-3.2-24B~\cite{mistralsmall32}, MedGemma-27B~\cite{medgemma}, and Claude Opus~5). Each model classified the frames using the same concept set without access to equipment annotations. An endorsement required agreement with the assigned concept at high or medium confidence, and at least three of six endorsements were required for retention. It removed $112$ of $543$ answerable items and the $31$ refusal items paired with their frames, yielding $581$ released items ($431$ answerable and $150$ refusal). Human ratings were collected separately to characterize annotation quality and did not modify released labels.

\subsection{Ethics statement}
\label{sec:ethics-statement}

The study used public TC3 videos and public United States Government and CoTCCC documents. No private clinical footage or patient-identifying material was used, no patients were enrolled, and no clinical intervention was performed. Four co-authors, two physicians and two medical students, rated the public video material. The raters were members of the research team assessing publicly available material, so the activity did not require institutional review.

\section{Data Records}
\label{sec:release}

TC3-VQA version 1.1 is deposited at Zenodo~\cite{tc3vqa2026data} (\url{https://doi.org/10.5281/zenodo.23170287}) and organized by content type. The data directory contains \texttt{items.jsonl}, with one record per item; \texttt{doctrine\_chunks.jsonl}, containing the cited corpus; \texttt{detection\_coco.json}, containing equipment boxes in COCO format; \texttt{segments.jsonl}, containing the $441$ source-video segments and their timestamps; and \texttt{frame\_features.npz}, containing image-encoder features for all frames. The meta directory provides the field dictionary (\texttt{FIELDS.md}), source-video manifest and license status, caption-mask rectangles, and excluded-item list. Of the $190$ video sources, $24$ were published by United States Government channels and one under a Creative Commons license; their $115$ frames from $75$ items are included in the frames directory. The remaining $165$ sources are represented by identifiers, URLs, timestamps, and annotations, with \texttt{scripts/fetch\_frames.py} provided to reconstruct the corresponding frames. Human adjudication is provided in the gold directory, the doctrine multiple-choice extension in the mcq directory, and evaluation inputs, controls, and scoring scripts in the eval directory. The release also includes a README, Croissant metadata (\texttt{croissant.json}), and a datasheet following the Datasheets for Datasets framework~\cite{gebru2021datasheets}.

\begin{figure*}[t]
\centering
\setlength{\tabcolsep}{2pt}
\begin{tabular}{cc}
\includegraphics[height=3.28cm]{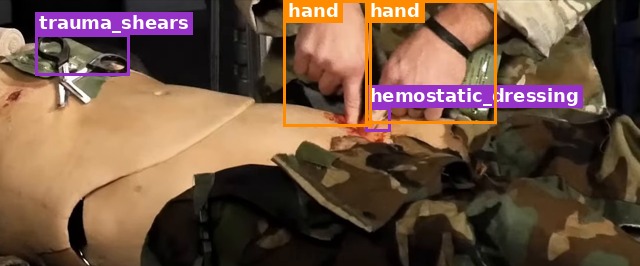} &
\includegraphics[height=3.28cm]{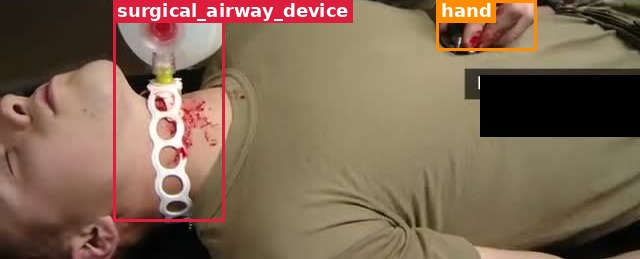} \\
{\small (a) Wound packing (M)} & {\small (b) Surgical cricothyroidotomy (A)} \\
\end{tabular}\\[4pt]
\begin{tabular}{ccc}
\includegraphics[height=2.56cm]{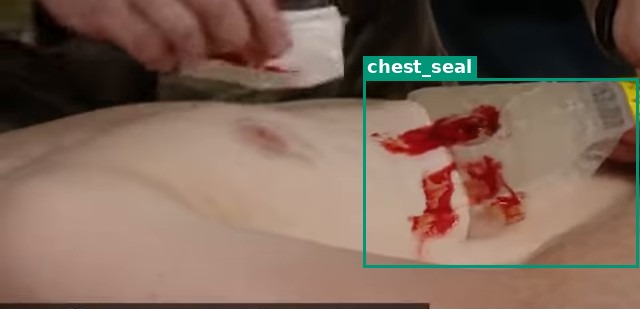} &
\includegraphics[height=2.56cm]{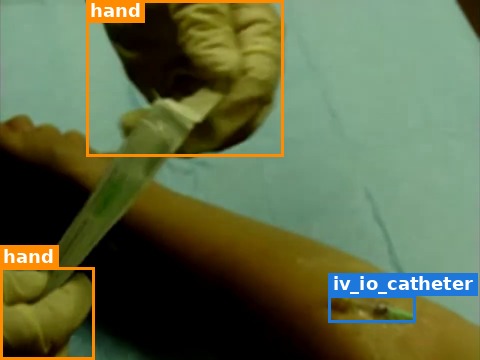} &
\includegraphics[height=2.56cm]{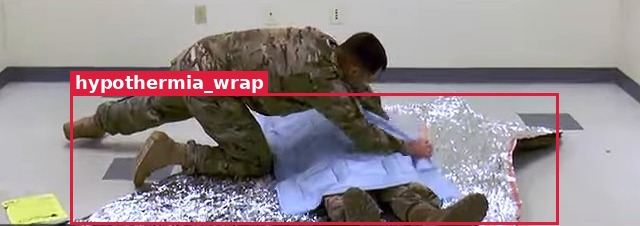} \\
{\small (c) Chest seal (R)} & {\small (d) IV/IO access (C)} & {\small (e) Hypothermia prevention (H)} \\
\end{tabular}
\caption{Example frames across the MARCH categories: (a) massive hemorrhage, (b) airway, (c) respiration, (d) circulation, and (e) hypothermia; IV/IO, intravenous or intraosseous. Frames are from public-domain CoTCCC footage. Overlays show released annotations from the $17$-class equipment ontology.}
\label{fig:gallery}
\end{figure*}

Each item links a frame or temporal window to its questions, reference answers, and visual annotations (Figure~\ref{fig:example}); \texttt{meta/FIELDS.md} defines all fields. Provenance fields identify the item and source video, URL and license status, video segment, and frame timestamps. Question fields record the type (recognition, doctrine, reasoning, procedural \textsc{how}, or refusal), text, recognition options, and reference answer. Doctrine, reasoning, and \textsc{how} answers include the source, corpus chunk, and \texttt{char\_start} and \texttt{char\_end} offsets of the cited span. Visual annotations include equipment boxes (17 classes), the answer region, and observed body region. Review fields include the assigned concept, safety-critical flag, visibility tier, audit results, and review-priority tier. Refusal items additionally record the missing-information category and the answerable item whose frames they reuse. Figure~\ref{fig:gallery} shows released annotations across the MARCH categories.

\section{Data Overview}
\label{sec:data-overview}

\begin{figure}[t]
\centering
\includegraphics[width=0.95\textwidth]{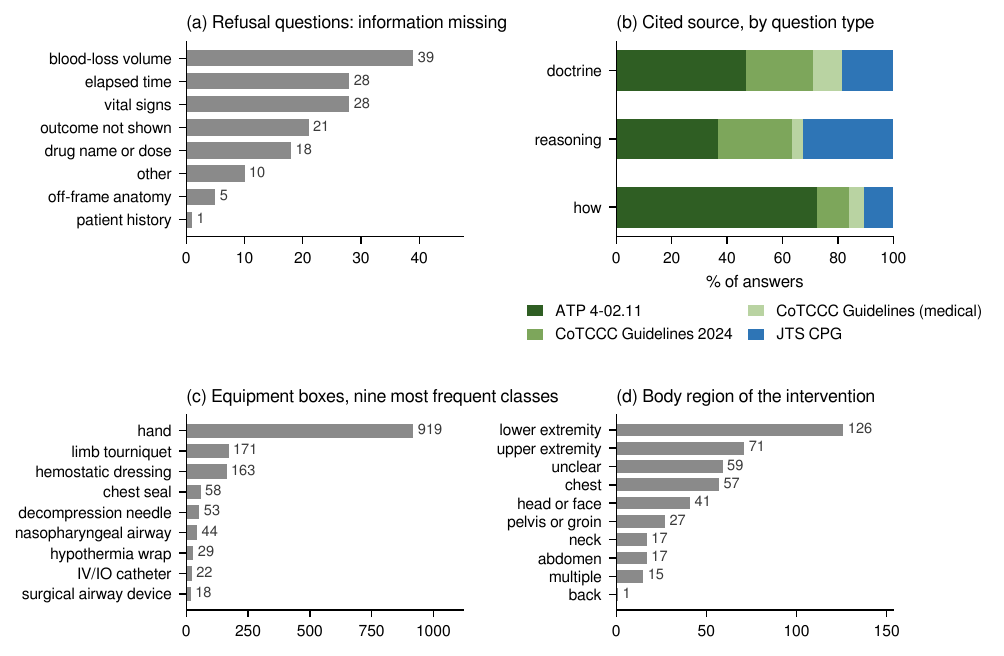}
\caption{Composition of the released layers. (a) Information requested by the $150$ refusal questions. (b) Source document of the cited answer, by question type. (c) Equipment-box counts for the nine most frequent of the $16$ classes present. (d) Observed body region of the $431$ answerable items.}
\label{fig:overview}
\end{figure}

TC3-VQA contains $1{,}860$ questions in $581$ items ($431$ answerable and $150$ refusal) drawn from $190$ videos and covering $11$ concepts (Table~\ref{tab:concepts}). The answerable set comprises $431$ recognition, $427$ doctrine, $425$ reasoning, and $427$ \textsc{how} questions, with all four types available for $418$ items. The sampled visual context is short, with a median segment duration of $10$\,s (interquartile range, $4$--$30$\,s). Of the $581$ items, $309$ use a single frame and $272$ use two to four frames. Each video contributes a median of three items and a maximum of eight, including refusals. Questions are one or two sentences long, with median lengths ranging from $9$ words for recognition to $27$ for refusal, while reference answers range from a $3$-word concept name to a $30$-word procedural passage. ATP~4-02.11 supplies approximately half of the cited answers and $73\%$ of procedural answers; the CoTCCC and Joint Trauma System guidelines supply most of the remainder (Figure~\ref{fig:overview}b). Refusal questions most often request blood-loss volume, elapsed time, or vital signs, with no category exceeding $26\%$ of refusal items (Figure~\ref{fig:overview}a). The release contains $1{,}565$ equipment boxes across $743$ of $903$ frames, covering $16$ of $17$ classes (Figure~\ref{fig:overview}c). Extremities are the most frequent intervention sites, followed by the chest (Figure~\ref{fig:overview}d).

\begin{table}[h]
\centering
\caption{Released items by concept and MARCH category. Safety-critical counts answerable items whose cited doctrine passage the stakes rubric flags as safety-critical; the risk-weighted hallucination rate (RWHR; Usage Notes) uses concept-level weights. The construction vocabulary also includes tourniquet conversion, which has no released item. NPA, nasopharyngeal airway; TXA, tranexamic acid.}
\label{tab:concepts}
\small
\begin{tabular}{l>{\raggedright\arraybackslash}p{5.4cm}rrr}
\toprule
MARCH & Concept & Answerable & Refusal & Safety-critical \\
\midrule
M (massive hemorrhage) & Wound packing & $129$ & $28$ & $129$ \\
                       & Limb tourniquet application & $137$ & $55$ & $137$ \\
                       & Junctional hemorrhage control & $4$   & $2$  & $4$ \\
\addlinespace
A (airway)             & Nasopharyngeal airway (NPA) insertion & $41$  & $17$ & $2$ \\
                       & Surgical cricothyroidotomy & $11$  & $5$  & $11$ \\
\addlinespace
R (respiration)        & Chest seal application & $29$  & $11$ & $29$ \\
                       & Needle chest decompression & $27$  & $8$  & $27$ \\
\addlinespace
C (circulation)        & IV/IO vascular access & $17$  & $9$  & $0$ \\
                       & TXA administration & $9$   & $3$  & $7$ \\
\addlinespace
H (hypothermia)        & Hypothermia prevention & $13$  & $6$  & $0$ \\
\addlinespace
process                & Casualty card (DD Form 1380) documentation & $14$  & $6$  & $0$ \\
\midrule
\multicolumn{2}{l}{Total} & $431$ & $150$ & $346$ \\
\bottomrule
\end{tabular}
\end{table}

\section{Technical Validation}
\label{sec:dataset_analysis}

\subsection{Concept labels}
\label{sec:audit}

Concept labels determine the questions associated with each scene, making their reliability central to the dataset. Labels were assessed at three stages: an independent visibility audit during construction (Sec.~\ref{sec:grounding}), cross-family consensus among six VLMs before release (Sec.~\ref{sec:verification}), and physician review of a stratified sample.

Among the $431$ released answerable items, $245$ were classified as clearly visible and $186$ as partly visible. Model consensus was unanimous for $147$ items, five of six for $117$, four for $91$, and three for $76$ (mean, $4.8$ of $6$). Individual-model agreement with the released label ranged from $57\%$ for MedGemma-27B to $94\%$ for Qwen2.5-VL-72B. Unanimous endorsement was more frequent for clearly visible items ($102/245$) than partly visible items ($45/186$). Physician review showed the same pattern: both physicians accepted $82\%$ of clearly visible labels compared with $44\%$ of partly visible labels, and acceptance decreased from $22/29$ items with unanimous model endorsement to $5/12$ items endorsed by three models (Sec.~\ref{sec:gradient}). Visibility and consensus strength are released with each item to support stratified evaluation.

\subsection{Question and answer validity}
\label{sec:release_audit}

Provenance was re-verified on the released records. All doctrine, reasoning, and \textsc{how} answers exactly reproduce their cited spans (Eq.~\ref{eq:offset}) and passed the entailment gate (Eq.~\ref{eq:nli}). We separately evaluated whether refusal questions were unanswerable from their frames, doctrine and \textsc{how} questions were grounded in their paired scenes, and reasoning answers correctly addressed their questions. Qwen3-VL-32B, InternVL3-78B, and MedGemma-27B independently judged every released item at temperature $0$, without access to the item type or concept label.

Each criterion included a negative control. Refusal questions were contrasted with answerable \textsc{how} questions from the same frames; doctrine and \textsc{how} questions were evaluated on their paired frames and on frames depicting another intervention; and reasoning answers were contrasted with answers from another item of the same concept. Table~\ref{tab:release_audit} reports criterion pass rates and corresponding control rates.

\begin{table}[h]
\centering\small\setlength{\tabcolsep}{4pt}
\caption{Automated audit of released questions and answers. Values are percentages satisfying each criterion; parentheses give the corresponding control rate. A majority requires two of three votes, and Gwet's AC1 measures inter-rater agreement. Anchored indicates that a question was judged specific to its paired frames.}
\label{tab:release_audit}
\begin{tabular}{lccccc}
\toprule
Judgment ($n$) & Qwen3-VL & InternVL3 & MedGemma & Majority & AC1 \\
\midrule
Refusal unanswerable ($150$) & $98.7$ ($12.2$) & $99.3$ ($34.5$) & $88.0$ ($3.6$) & $\mathbf{98.7}$ ($10.8$) & $0.91$ \\
Doctrine question anchored ($427$) & $53.9$ ($4.7$) & $50.4$ ($6.6$) & $63.5$ ($52.7$) & $\mathbf{58.8}$ ($7.7$) & $0.46$ \\
\textsc{how} question anchored ($427$) & $87.8$ ($4.9$) & $93.4$ ($9.1$) & $94.1$ ($89.2$) & $\mathbf{94.6}$ ($11.0$) & $0.86$ \\
Reasoning answer correct ($425$) & $86.4$ ($30.8$) & $89.4$ ($33.2$) & $90.8$ ($38.0$) & $\mathbf{91.5}$ ($33.0$) & $0.84$ \\
\bottomrule
\end{tabular}
\end{table}

The majority judged $98.7\%$ of refusal questions unanswerable, compared with $10.8\%$ of answerable controls on the same frames, and $94.6\%$ of \textsc{how} questions were anchored to their paired frames, compared with $11.0\%$ on substituted frames. Doctrine questions were less scene-specific ($58.8\%$), reflecting that $37\%$ asked about indications or guidance applicable across scenes of the same intervention. MedGemma-27B showed limited sensitivity to the \textsc{how} control, judging $89.2\%$ as anchored even on substituted frames.

The reasoning audit accepted $91.5\%$ of released answers compared with $33.0\%$ of within-concept controls. It accepted $48$ of the $49$ answers accepted by both physicians, but also $35$ of the $38$ flagged by at least one physician, indicating lower sensitivity to deficiencies identified by expert review. Because the same model families also contributed to concept-label validation, these results provide a model-based consistency check; clinical adequacy is evaluated separately by the physician ratings below.

\subsection{Equipment annotations}
\label{sec:detection-validation}

The release contains $1{,}565$ boxes across $743$ of $903$ frames, covering $16$ of the $17$ ontology classes. Among the $431$ answerable items, $383$ contain equipment boxes, $393$ contain answer-region boxes, and $420$ contain at least one of the two. Every released box was individually reviewed during construction (Sec.~\ref{sec:grounding}) to verify object presence and class assignment. Because the SAM-refined coordinates were not modified during review, this verification establishes box content rather than localization accuracy.

As a consistency check, a YOLOv8s detector~\cite{ultralytics} trained on $595$ frames and evaluated on $148$ held-out frames achieved $\mathrm{mAP}_{50}=0.49$ and $\mathrm{mAP}_{50\text{-}95}=0.35$. Class-level $\mathrm{mAP}_{50}$ was approximately $0.6$ for hands, decompression needles, hypothermia wraps, and surgical-airway devices, compared with $0.28$ for the more visually variable hemostatic dressings. Because the split was defined by frame rather than source video, these results characterize annotation consistency within the collection rather than generalization to unseen footage.

\subsection{Human ratings}
\label{sec:gradient}

Two physicians (one TC3-trained) and two medical students independently evaluated $120$ candidate items sampled across concepts and visibility tiers, with at most two items per distinct doctrine answer. Subsequent release filtering retained $88$ of these items, which form the basis of the results below. Using a prespecified rubric, raters assessed recognition labels as correct, unclear, or wrong; the three quoted answers as correct, correct with a minor problem, or incorrect; and the clinical severity of errors. Agreement is reported using Gwet's AC1 because ratings were strongly skewed toward correct.

\begin{table}[h]
\centering
\caption{Human ratings on the $88$ adjudicated items retained in the release. Cells report correct / unclear / wrong for recognition and correct / minor problem / incorrect for quoted answers. AC1 is Gwet's agreement within each rater pair and among all four raters.}
\label{tab:gradient}
\small
\renewcommand{\arraystretch}{1.5}
\setlength{\tabcolsep}{5pt}
\begin{tabular*}{\linewidth}{@{\extracolsep{\fill}}lcccc@{\hspace{10pt}}ccc@{}}
\toprule
 & \multicolumn{2}{c}{Physicians (\%)} & \multicolumn{2}{c}{Medical students (\%)} & \multicolumn{3}{c}{Agreement (AC1)} \\
\cmidrule(lr){2-3}\cmidrule(lr){4-5}\cmidrule(l){6-8}
Question type & Rater 1 & Rater 2$^{\dagger}$ & Rater 1 & Rater 2 & Physicians & Students & All four \\
\midrule
Recognition & $69/20/10$ & $69/23/8$ & $93/3/3$ & $94/1/5$ & $0.80$ & $0.95$ & $0.75$ \\
Doctrine & $74/17/9$ & $91/3/6$ & $95/3/1$ & $76/17/7$ & $0.66$ & $0.71$ & $0.68$ \\
Reasoning & $59/30/11$ & $92/1/7$ & $92/6/2$ & $60/33/7$ & $0.48$ & $0.52$ & $0.57$ \\
Procedural (\textsc{how}) & $68/28/3$ & $93/6/1$ & $93/7/0$ & $41/50/9$ & $0.61$ & $0.26$ & $0.47$ \\
\bottomrule
\end{tabular*}
\par\smallskip
\begin{minipage}{\linewidth}
\footnotesize $^{\dagger}$TC3-trained physician. Rows may not sum to $100$ because of rounding and up to two skipped ratings. Individual ratings and notes are provided in \texttt{gold/}.
\end{minipage}
\end{table}

Both physicians rated $69\%$ of recognition labels correct, with AC1 $0.80$; uncertainty was more common than outright rejection (Table~\ref{tab:gradient}). The medical students accepted more labels ($93$--$94\%$ correct), including $23$ of the $30$ questioned by at least one physician. Physician uncertainty occurred mainly in partly visible scenes and when adjacent procedural steps, visually similar devices, or unreadable details made the assigned intervention ambiguous. These items were also harder for the five baseline models, whose mean recognition accuracy was $0.79$ on labels accepted by both physicians and $0.65$ on those questioned by both (difference $0.14$, bootstrap $95\%$ confidence interval $[0.00,0.28]$). The released ratings therefore provide an expert-based indicator of item difficulty rather than a definitive relabeling of the dataset.

Across the three quoted-answer types, outright incorrect ratings ranged from $0$ to $11\%$, while differences between raters largely reflected use of the minor-problem category. Notes most often identified context omitted by the quoted passage and occasionally challenged the source itself. Severity ratings showed lower physician agreement (AC1 $0.27$), although mean physician severity increased from $1.0$ to $2.05$ and $2.4$ across concepts assigned RWHR weights of $1$, $2$, and $3$, fixed before rating, respectively ($n=7$, $10$, and $71$).

\subsection{Baseline performance and image dependence}
\label{sec:results}
\label{sec:controls}

We benchmarked five open VLMs with $7$--$38$ billion parameters: Qwen2-VL-7B~\cite{qwen2vl}, Qwen2.5-VL-7B, InternVL3-8B and 38B, and Phi-3.5-Vision~\cite{phi3}. Models used for primary perception and question generation were excluded, although Qwen and InternVL families contributed to other construction stages. Recognition and refusal were evaluated as five-option questions with abstention. Doctrine was evaluated by free-form entailment and by multiple choice using distractors from another concept (easy) or another facet of the same concept (hard). RWHR (Sec.~\ref{sec:rwhr}) measures severity-weighted error among answered items.

\begin{table}[h]
\centering
\caption{Baseline performance on $431$ answerable and $150$ refusal items. Doctrine-MCQ (multiple-choice questions) uses distractors from other concepts (easy) or other facets of the same concept (hard). Doctrine-open reports reference-sentence entailment and the proportion of response sentences unsupported by the reference or cited passage. Bold indicates the best value in each column; lower is better for unsupported and RWHR.}
\label{tab:results}
\begin{tabular}{lccccccc}
\toprule
Model & Recognition & \multicolumn{2}{c}{Doctrine-MCQ} & \multicolumn{2}{c}{Doctrine-open} & Refusal & RWHR~$\downarrow$ \\
 & & easy & hard & entailed & unsupported~$\downarrow$ & & \\
\midrule
Qwen2-VL-7B     & $\mathbf{0.854}$ & $0.951$ & $0.864$ & $0.267$ & $0.576$ & $0.260$ & $0.921$ \\
Qwen2.5-VL-7B   & $0.805$ & $0.963$ & $0.874$ & $\mathbf{0.285}$ & $0.574$ & $0.833$ & $0.546$ \\
Phi-3.5-Vision  & $0.666$ & $0.867$ & $0.703$ & $0.225$ & $0.588$ & $\mathbf{0.933}$ & $0.888$ \\
InternVL3-8B    & $0.840$ & $0.948$ & $0.845$ & $0.243$ & $0.579$ & $0.860$ & $0.479$ \\
InternVL3-38B   & $0.740$ & $\mathbf{0.986}$ & $\mathbf{0.913}$ & $0.264$ & $\mathbf{0.523}$ & $\mathbf{0.933}$ & $\mathbf{0.420}$ \\
\bottomrule
\end{tabular}
\end{table}

\begin{figure}[h]
\centering
\includegraphics[width=0.8\textwidth]{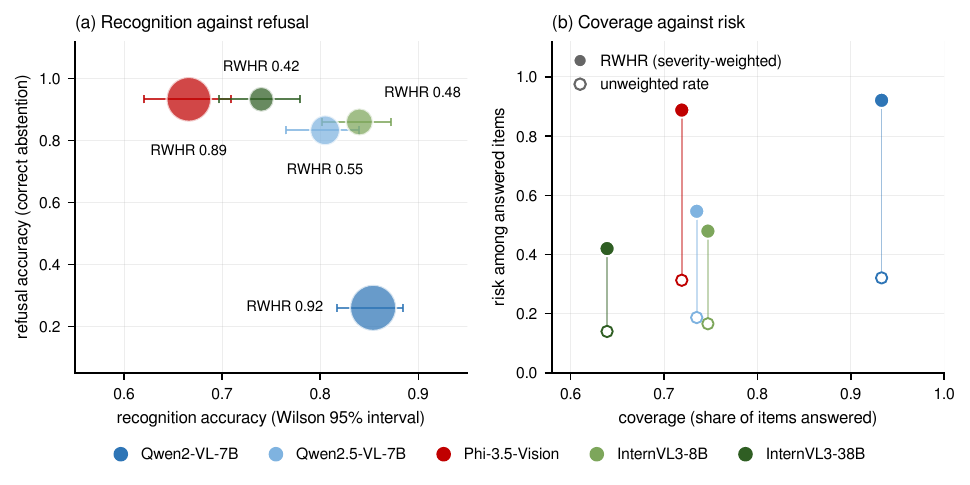}
\caption{Recognition, abstention, and error across five baseline models. (a) Recognition and refusal accuracy with Wilson $95\%$ intervals; marker area represents RWHR. (b) Coverage versus unweighted error (open markers) and severity-weighted error (filled markers).}
\label{fig:baselines}
\end{figure}

Recognition accuracy ranged from $0.67$ to $0.85$ (Table~\ref{tab:results}), with lower performance on partly visible interventions. Recognition and refusal showed a clear trade-off: Qwen2-VL-7B achieved the highest recognition accuracy ($0.85$) but only $26\%$ refusal accuracy and the highest RWHR ($0.92$), whereas InternVL3-38B achieved $0.74$, $93\%$, and $0.42$, respectively (Figure~\ref{fig:baselines}). Doctrine-MCQ accuracy ranged from $0.87$ to $0.99$ with easy distractors and from $0.70$ to $0.91$ with hard distractors. Free-form responses entailed $0.23$--$0.29$ of reference sentences, while $52$--$59\%$ of generated sentences were unsupported.

Recognition depended strongly on the visual input. Substituting frames from another intervention reduced accuracy to $0.04$--$0.14$ for four models, below the $0.20$ uniform-choice reference, while Phi-3.5-Vision retained $0.33$ (Table~\ref{tab:controls}). Multi-frame context also contributed information: restricting the $204$ multi-frame items to the first frame reduced accuracy by $7.8$--$12.7$ percentage points for four models, with significant exact McNemar tests, while Qwen2-VL-7B decreased by $2.5$ points ($p=0.42$). Full-window reruns reproduced the released scores within $0.002$.

\begin{table}[h]
\centering\small
\caption{Recognition under frame substitution and temporal truncation. Left, questions paired with frames from another concept ($431$ items; uniform-choice reference $0.20$). Right, full windows compared with the first frame alone ($204$ multi-frame items). $\Delta$ denotes perturbed minus original accuracy; $p$ values are from two-sided exact McNemar tests.}
\label{tab:controls}
\begin{tabular}{lcccc@{\hskip 18pt}cccc}
\toprule
 & \multicolumn{4}{c}{Frame substituted (all $431$)} & \multicolumn{4}{c}{First frame only ($204$ multi-frame)} \\
\cmidrule(lr){2-5}\cmidrule(lr){6-9}
Model & own & swapped & $\Delta$ & abstain & window & $1$ frame & $\Delta$ & $p$ \\
\midrule
Qwen2-VL-7B   & $0.854$ & $0.139$ & $-0.715$ & $2.3\%$ & $0.853$ & $0.828$ & $-0.025$ & $0.42$ \\
Qwen2.5-VL-7B & $0.807$ & $0.081$ & $-0.726$ & $18.8\%$ & $0.814$ & $0.686$ & $-0.127$ & $<0.001$ \\
Phi-3.5-Vision& $0.666$ & $0.332$ & $-0.334$ & $10.2\%$ & $0.716$ & $0.637$ & $-0.078$ & $0.014$ \\
InternVL3-8B  & $0.840$ & $0.123$ & $-0.717$ & $17.9\%$ & $0.873$ & $0.745$ & $-0.127$ & $<0.001$ \\
InternVL3-38B & $0.740$ & $0.044$ & $-0.696$ & $44.8\%$ & $0.819$ & $0.725$ & $-0.093$ & $0.002$ \\
\bottomrule
\end{tabular}
\end{table}

\subsection{Comparison with direct question-answer generation}
\label{sec:conventional}

We compared the construction pipeline with direct question-answer generation on $288$ candidates whose assigned concepts were independently confirmed as clearly visible. Six open VLMs and Claude Opus~5 generated one pair for each answerable question type from the same frames, with answer length matched to the reference and without access to the audited concept, doctrine corpus, or verification gates. Qwen2.5-VL-72B was additionally tested with the audited concept provided to isolate the effect of concept verification.

\begin{figure*}[h]
\centering
\includegraphics[width=\textwidth]{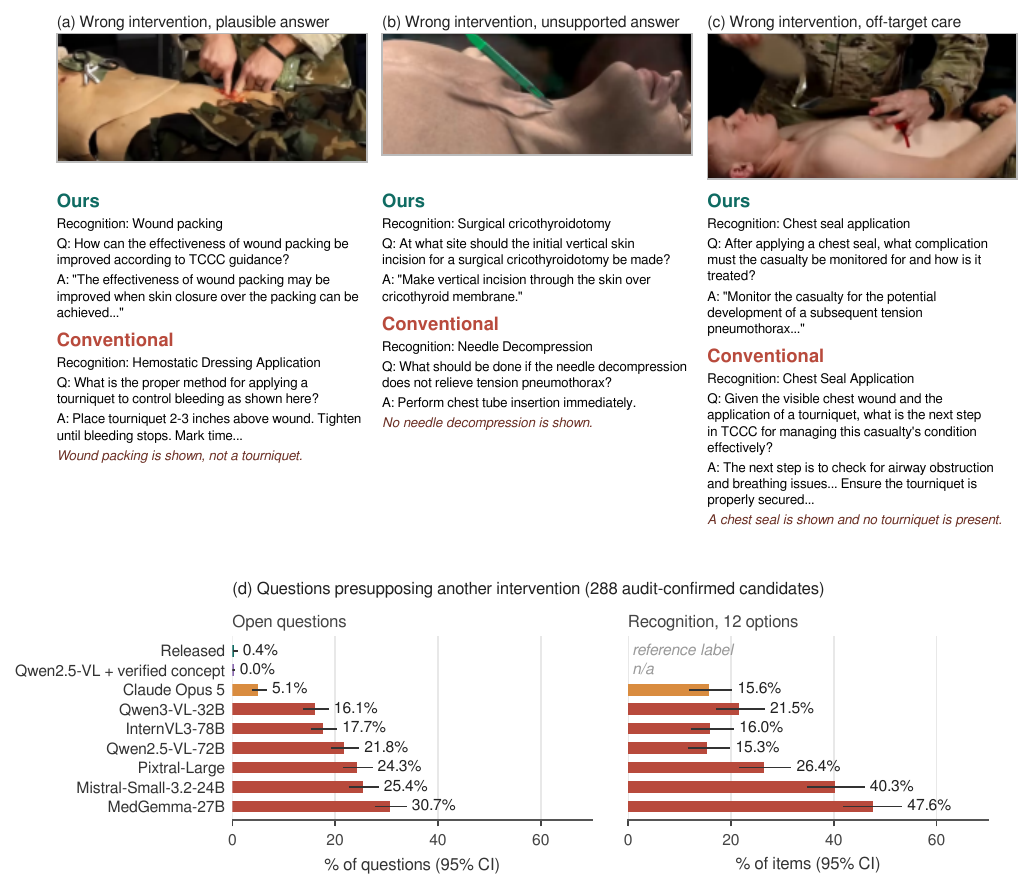}
\caption{Reference question-answer (QA) pairs compared with direct generation from the same frames. (a--c) Public-domain examples showing the pipeline-generated reference pair (Ours) and direct generation by Qwen2.5-VL-72B (Conventional). (d) Concept inconsistency during direct question generation (left) and closed-set concept selection (right) on $288$ audit-confirmed candidates (Pixtral-Large~\cite{pixtrallarge} and the models of Sec.~\ref{sec:verification}), with Wilson $95\%$ intervals.}
\label{fig:conventional}
\end{figure*}

Direct generation frequently produced questions about the wrong intervention. Among the open VLMs, $16$--$31\%$ of questions named another listed concept, compared with $0.4\%$ of pipeline-generated questions. When models selected the concept themselves before generation, $11$--$42\%$ of questions remained inconsistent with the audited concept, whereas Qwen2.5-VL-72B produced no such errors when the audited concept was provided (Figure~\ref{fig:conventional}d).

Five blinded large language model (LLM) judges from different model families (Llama-3.3-70B~\cite{llama33}, MedGemma-27B, Phi-4~\cite{phi4}, DeepSeek-R1-Distill-Llama-70B~\cite{deepseekr1}, and gpt-oss-120b~\cite{gptoss}) assessed whether each question was concept-consistent and its answer supported by eight doctrine passages retrieved for that question. Pipeline-generated pairs satisfied both criteria for $95$--$99\%$ of questions, compared with $72$--$78\%$ for direct generation by Qwen2.5-VL-72B. Among concept-consistent questions, however, at least $92\%$ of answers in both groups satisfied both criteria under every judge. The difference therefore arose primarily from concept misalignment during question generation rather than unsupported answers once the intervention was correctly identified.

\FloatBarrier

\section{Usage Notes}
\label{sec:eval}

\subsection{Model adaptation and data partitioning}

For instruction tuning, pair the referenced frames with each question and released answer, preserving source citations and audit metadata. Refusal examples provide contrasting supervision from the same scenes, where an intervention may be recognizable even when the requested clinical detail is unsupported.

Training, validation, and test sets should be partitioned by source video, with each answerable item kept with its derived refusals to prevent frame overlap. Because $1{,}279$ quoted answers cite only $313$ distinct passages from $49$ chunks across six documents, studies of transfer to unseen doctrine should additionally separate source passages and report concept coverage. Reports should specify the dataset version, included question types, and use of human ratings.

\subsection{Recognition and abstention}

Recognition is evaluated by top-1 accuracy, with macro accuracy over the six MARCH categories reported for the imbalanced concept distribution:
\begin{equation}
\mathrm{Acc}_{\text{macro}} \;=\; \frac{1}{K}\sum_{k=1}^{K} \frac{1}{|I_k|}\sum_{i \in I_k} \mathbf{1}\!\left[\hat{y}_i = y_i\right],
\label{eq:macro}
\end{equation}
where $I_k$ denotes items in category $k$ and $K=6$. Refusal performance is the proportion of unanswerable items for which the model abstains and should be reported separately from recognition accuracy.

Results can also be stratified by intervention visibility, consensus strength, concept frequency, frame count, and, for the adjudicated subset, physician acceptance. Each answerable item additionally carries a safety-critical flag for its cited doctrine passage, distinct from the concept-level severity weight used by RWHR.

\subsection{Doctrine scoring}

Free-form doctrine answers are scored by entailment against the reference answer and cited passage using the NLI classifier and threshold of Eq.~\ref{eq:nli}; the multiple-choice variant is scored by accuracy. Sentence-level coverage is
\begin{equation}
\mathrm{Doctrine}_{\text{partial}}(\hat{r}, g) \;=\; \frac{1}{|\mathrm{sent}(g)|} \sum_{u \in \mathrm{sent}(g)} \mathbf{1}\!\left[\Pr\nolimits_{\textsc{nli}}\!\left(\text{entail} \mid \hat{r}, u\right) \ge \tau\right],
\label{eq:doctrine}
\end{equation}
with the response as premise and each reference sentence as hypothesis. Whole-answer entailment is also reported. Unsupported content is measured as the proportion of response sentences entailed by neither the reference nor the cited passage and should be interpreted cautiously because the classifier may miss valid paraphrases.

Doctrine-MCQ uses the cited answer as the reference option. Easy distractors come from the same facet of another concept, whereas hard distractors come from another facet of the same concept; options are filtered for near-duplicates and randomized. Both variants should be reported alongside free-form scores because they test different response requirements.

\subsection{Visual grounding annotations}

Equipment boxes are provided in COCO format over the $17$-class ontology, together with an answer region and observed body region for each answerable item. These annotations support equipment detection, evidence localization, and analyses of model attention to relevant instruments and anatomy. Box coverage is not exhaustive, so missing boxes should not be treated as negative annotations. Detection experiments should split data by source video.

\subsection{Risk-weighted error and coverage}
\label{sec:rwhr}

The Risk-Weighted Hallucination Rate (RWHR) measures severity-weighted recognition errors and responses to unanswerable questions among non-abstained items. Concepts receive weights $w\in\{1,2,3\}$ for minor, serious, and critical consequences of error: $3$ for hemorrhage, airway, and respiration concepts; $2$ for intravenous or intraosseous access, tranexamic acid, and hypothermia prevention; and $1$ for documentation. Let $a_i$ indicate a non-abstained response, $g_i$ the reference label, and $\varnothing$ abstention:
\begin{equation}
\mathrm{RWHR} \;=\; \frac{\sum_i a_i w_i h_i}{\sum_i a_i},
\qquad
h_i = \mathbf{1}\!\left[\hat{y}_i\neq\varnothing \;\wedge\; \hat{y}_i\neq g_i\right].
\end{equation}

RWHR is an average weighted error per answered item and can exceed one; it is undefined at zero coverage. It should therefore be reported with coverage, unweighted error, and the risk--coverage curve. Effective Reliability provides a complementary summary~\cite{whitehead2022reliability}. The weighting scheme is dataset-specific and should not replace unweighted results.

\subsection{Limitations}
\label{sec:limitations}

TC3-VQA contains automatically selected reference labels, with human ratings available for a stratified subset. Exact citations establish provenance, but automated verification can preserve shared perceptual or textual errors, and the $88$ retained adjudicated items do not establish clinical accuracy for every record. Human ratings and audit metadata should therefore inform uses requiring higher annotation confidence.

Coverage reflects the available footage and doctrine. Tourniquet application and wound packing dominate, while sparse concepts may require reporting by MARCH category. Some visual details, including medication identity, dose, route, and timing, cannot reliably be resolved from sampled frames, and repeated doctrine passages may make some questions answerable from general knowledge once the intervention is known.

The consensus includes model families related to several baselines, and detector validation used a frame-level split that may share source-video characteristics. Instructional and field-training footage also differs from real combat casualties; performance on TC3-VQA therefore does not establish deployment readiness. Broader clinical adjudication, clinician-corrected labels, and transfer to operational settings remain future work.

\section{Data Availability}
\label{sec:data-availability}
TC3-VQA version 1.1 is deposited at Zenodo, \url{https://doi.org/10.5281/zenodo.23170287}~\cite{tc3vqa2026data}. The package includes item records, the cited doctrine chunks, visual annotations, source and license manifests, temporal segments, caption masks, the human ratings, doctrine multiple-choice tasks, evaluation inputs and scorers, and the results reported here. Image files are included for $115$ frames from $25$ sources with redistribution permission; the remaining frames are specified by source identifier and timestamp and are regenerated from the original videos with the script provided. For items whose pixels cannot be redistributed, the package also carries frozen image-encoder features of every frame, so all $581$ items remain usable without the source videos. Source reachability was checked on 17 September 2026: $189$ of the $190$ videos were reachable, and one, covering two items, had been removed. Annotations are released under CC BY 4.0 and the scripts under the MIT license. Construction artifacts are retained by the authors and available on request.

\section{Code Availability}
\label{sec:repro}

The construction and evaluation code is available at \url{https://github.com/Kamaleswaran-Lab/TC3-VQA}. It includes frame reconstruction, annotation, retrieval, verification, and scoring. Qwen2.5-VL-72B-Instruct~\cite{qwen25vl} was served with vLLM 0.15 and tensor parallelism of four on NVIDIA H200 (141\,GB) GPUs. Images were resized to a longest side of at most $896$ pixels. Claude Opus~4.8 performed visual auditing, doctrine-facet mining, and review.
\bibliographystyle{unsrt}
\bibliography{references}

\section*{Author Contributions}
J.K.: Conceptualization, Methodology, Software, Validation, Formal analysis, Investigation, Data curation, Visualization, Writing -- original draft, Writing -- review \& editing; J.C.: Software, Data curation, Writing -- review \& editing; A.A.: Validation, Writing -- review \& editing; V.M.: Validation, Writing -- review \& editing; Y.L.: Validation, Writing -- review \& editing; P.-C.C.: Validation, Writing -- review \& editing; S.H.: Methodology, Writing -- review \& editing; R.K.: Supervision, Funding acquisition, Writing -- review \& editing.

\section*{Competing Interests}
The authors declare no competing interests.

\section*{Funding}
This work was supported in part by the National Institutes of Health under Grant R01GM139967 and Grant R01HL170175.

\end{document}